\documentclass[conference]{IEEEtran}

\IEEEoverridecommandlockouts
\usepackage{graphicx}
\usepackage{subfig}
\usepackage[T1]{fontenc}
\usepackage{hyperref}
\graphicspath{ {./Images/} }
\begin{document}

\title{Using an Uncrewed Surface Vehicle to Create a Volumetric Model of Non-Navigable Rivers and Other Shallow Bodies of Water}

\author{\IEEEauthorblockN{Jayesh Tripathi}\IEEEauthorblockN{Robin R. Murphy~\IEEEmembership{Fellow,~IEEE}}
\thanks{J. Tripathi and R. Murphy are with Texas A\&M University, College Station, TX \{jtjayesh98,robin.r.murphy\}@tamu.edu}
\thanks{This work has been supported through a subcontract to Hydronalix for DOD-Naval Air Warfare Center Aircraft Division STTR 140D0420C0097. The contents of this article do not constitute an endorsement by either NAWCAD or Hydronalix.}}

\maketitle

\begin{abstract}
Non-navigable rivers and retention ponds play important roles in buffering communities from flooding, yet emergency planners often have no data as to the volume of water that they can carry before flooding the surrounding.
This paper describes a practical approach for using an uncrewed marine surface vehicle (USV) to collect and merge bathymetric maps with digital surface maps of the banks of shallow bodies of water into a unified volumetric model.  
The below-waterline mesh is developed by applying the Poisson surface reconstruction algorithm to the sparse sonar depth readings of the underwater surface. Dense above-waterline meshes of the banks are created using commercial structure from motion (SfM) packages.
Merging is challenging for many reasons, the most significant is gaps in sensor coverage, i.e., the USV cannot collect sonar depth data or visually see sandy beaches leading to a bank thus the two meshes may not intersect. 
 The approach is demonstrated on a Hydronalix EMILY USV with a Humminbird single beam echosounder and Teledyne FLIR camera at Lake ESTI at the Texas A\&M Engineering Extension Service Disaster City\textsuperscript{\textregistered} complex. 
  
\end{abstract}

\section{Introduction}

Prediction of floods is hampered by a low-cost system to generate volumetric models of non-navigable rivers, retention ponds, and other small bodies of water. The lack of volumetric models means that emergency managers cannot estimate the carrying or holding capacity of the body of water or at what point rainfall would result in a flood. There has been considerable work in using small uncrewed marine surface vehicles (USV) with low-cost depth finders to conduct bathymetric mapping \cite{sotelo-torres:2023,nguyen:2022,li:2021,Sprech:2020,Kurowski:2019,Cezary:2017,Naing:2017,Metcalfe:2017,Bakar:2017}. Unfortunately, a bathymetric depth profile is not sufficient because the elevation profile of the banks of the body of water is needed to complete a useful 3D volumetric model.

One approach to acquiring the above-waterline elevation data is to use an uncrewed aerial system (UAS). However, a low-cost UAS using only a camera may not be able to generate a useful digital elevation map (DEM) of  the banks due to the presence of trees, piers, etc. 
An additional disadvantage is that the addition of a UAS increases the workforce to field both a USV and UAS for the same area. 

\begin{figure}[h]
\centering
\includegraphics[width=0.47\textwidth]{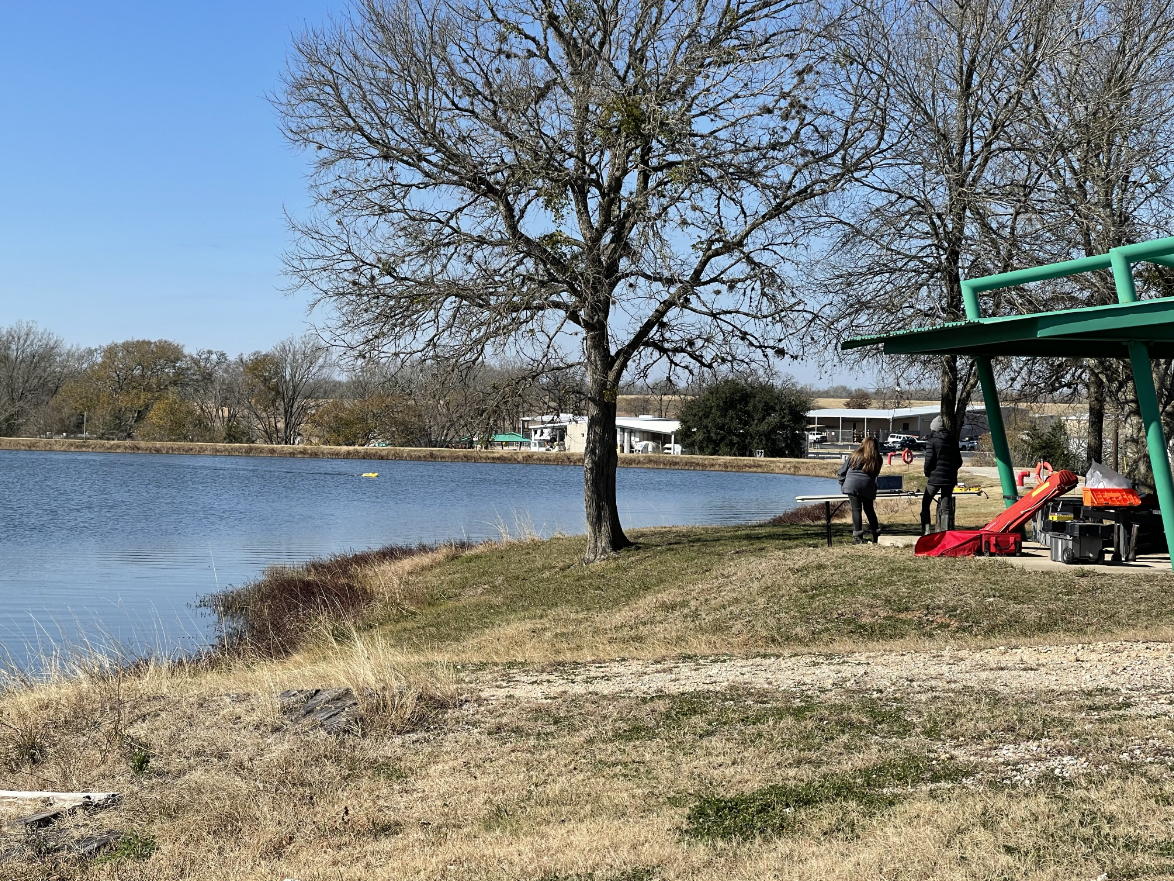}
\caption{EMILY USV surveying Lake ESTI.}
\label{fig:emily}
\end{figure}

Our approach, similar to Mancini et. al \cite{Mancini:2015, ManciniUAS:2015}, is to use a USV with sonar for bathymetry plus a camera mounted on top to collect the above-waterline imagery for the creation of a DEM. 
This approach goes further than \cite{Mancini:2015} by merging the two 3D meshes into a single 3D mesh which can then be imported into geographical information system packages.  This method is low-cost, relying on open-source software whenever possible and minimizes the workforce needed. 
It was demonstrated on a Hydronalix EMILY USV \cite{Patterson:2012} with a Humminbird single beam echosounder and a Teledyne FLIR Duo Pro R camera for a small pond (See Fig.~\ref{fig:emily}). The approach is expected to extend to other applications such as detecting changes over time in sediment deposits\cite{rangel:2019} and erosion in such bodies of water\cite{Bio:2015}, and glacial lake outbursts\cite{li:2021}.


\section{Prior Work}

Prior work in designing and using a USV to collect a complete volumetric model of shallow bodies of water appears extremely limited.
McLaren Engineering \cite{mclaren}, appears to use small USV and UAS in generating digital reconstruction of marine sea beds and coastal structures. 
However, the robots appear to use expensive multibeam sonar and Lidar rather than single-beam sonar
and there is no indication of whether the digital surveys are merged and if so how much of the process is automated. 
An Italian team Mancini et. al \cite{Mancini:2015} demonstrated a small USV with a SonarMite v3 Echo Sounder for bathymetric data collection and a GoPro Hero for above-waterline data, and a real-time kinematic (RTK) positioning system for GPS. The team experimented with a Kinect V2 depth camera but river banks with trees generated noisy data \cite{ManciniUAS:2015}. The imagery from the GoPro was manually edited to remove the water and sky, and then Agisoft Photoscan was used to generate a DSM.  The two data sets were not fused into a unified model.

\section{Approach}

\begin{figure}[h]
\centering
\includegraphics[width=0.45\textwidth]{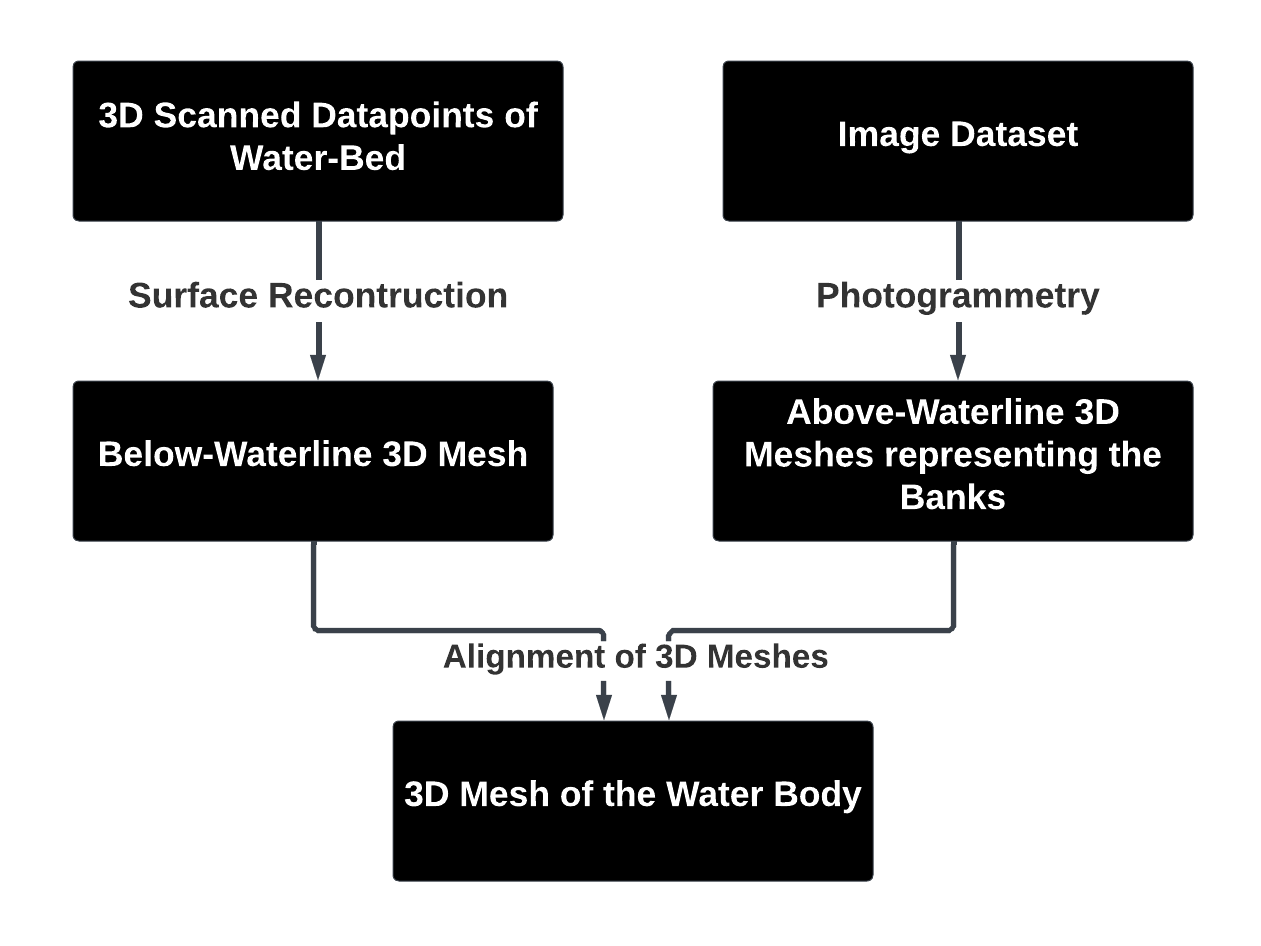}
\caption{Overview on How to Generate 3D Mesh.}
\label{fig:overview}
\end{figure}

The approach taken by this project was to use a single USV configured with both a depth finder, camera, and GPS with RTK, similar to \cite{Mancini:2015}. As shown in Fig.~\ref{fig:overview},
the depth finder produces a sparse bathymetric 3D mesh, while the camera produces a DSM from structure from motion which is converted to a 3D mesh. The conversion of both below- and above-waterline readings into a 3D mesh provides approximate surfaces which are more amenable to
 volumetric reasoning. The two meshes are then aligned into a single 3D mesh.

\subsection{Generating the Below-Waterline 3D Mesh}



\begin{figure}[h]
\centering
\includegraphics[width=0.35\textwidth]{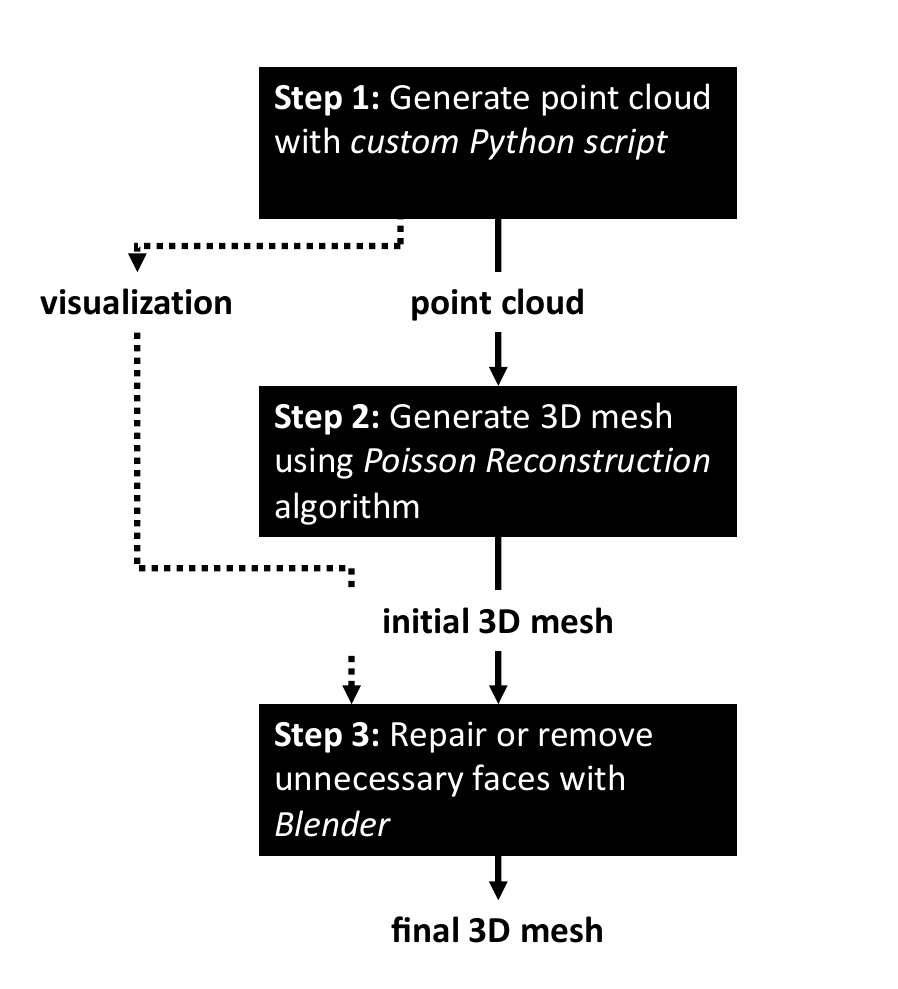}
\caption{Steps in generating the Below-Waterline Mesh}
\label{fig:OverviewBWL}
\end{figure}

\begin{figure}[h]
\centering
\includegraphics[width=0.4\textwidth]{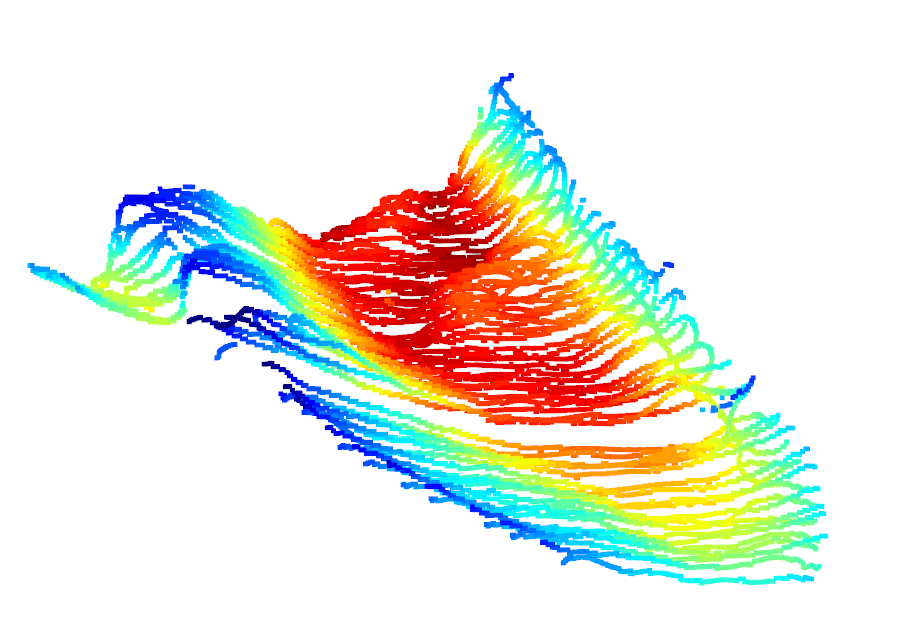}
\caption{Visualization of the depth readings.}
\label{fig:BWLpre}
\end{figure}

As shown in Fig.~\ref{fig:OverviewBWL}, the process of generating the below-waterline 3D mesh has three steps. 
In Step 1, the USV performs a complete coverage survey of the body of water \cite{trey}, producing a .txt file of depth readings with the associated GPS latitude and longitude, essentially a sparse point cloud. The Humminbird data reports latitude and longitude in degrees (measured in WGS84) and depth  in meters. The latitude and longitude are converted into the UTM Zone projection, which measures each coordinate in meters, to maintain uniformity. It is not necessary but recommended, to visualize the depth readings as a depth map, see Fig.~\ref{fig:BWLpre}; that visualization serves to verify the correctness of the 3D map in Step 3. The custom script also estimates the centre of the convex hull of the depth map to simplify locating the mesh when working with Blender in Step 3. 

 In Step 2,  the Python script applies the Open3D Library Poisson Reconstruction algorithm\cite{Open3Dlibary} to create the 3D mesh. Note that the Open3D Ball Pivoting algorithms were experimented with. The Ball Pivoting algorithm was discarded because it did not generate meshes that matched the depth map. In addition, the Ball Pivoting algorithm had more parameters to tune, while the Poisson was more straightforward for a partially automated workflow. 

\begin{figure}[!tbp]
  \centering
  \subfloat[Unrepaired Mesh. The highlighted regions in orange represent the unwanted faces produced during the meshing process)]{\includegraphics[width=0.45\textwidth]{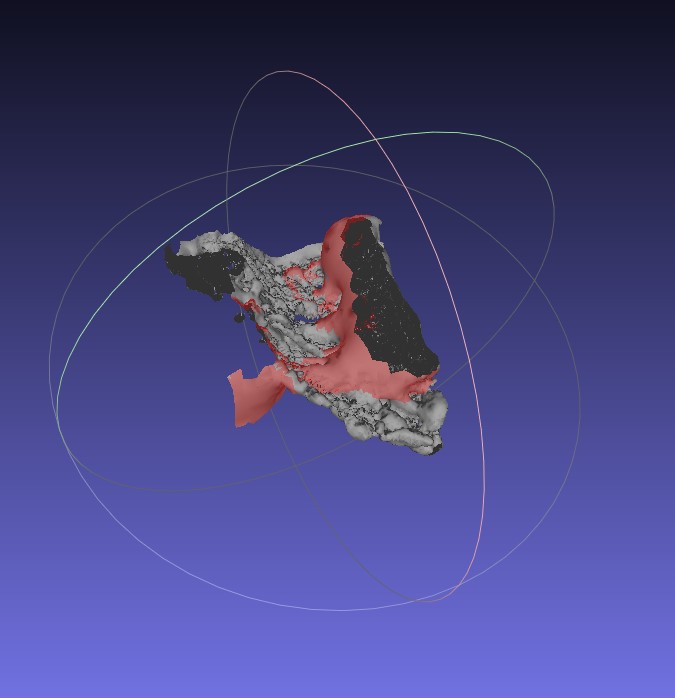}\label{fig:f1}}
  \hfill
  \subfloat[Repaired 3D Mesh.]{\includegraphics[width=0.45\textwidth]{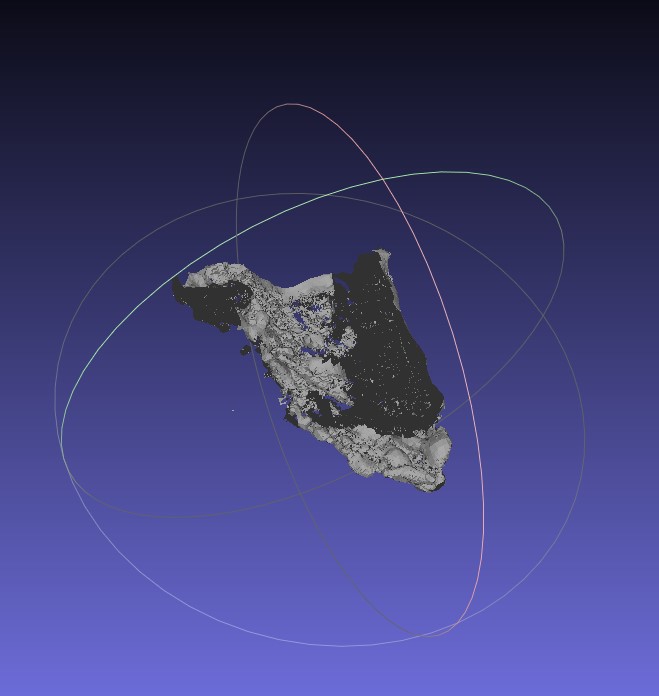}\label{fig:f2}}
  \caption{Example of Step 3, repairing 3D meshes by removing unnecessary faces.}
  \label{fig:BWLmesh}
\end{figure}

Step 3 is to repair or remove unnecessary faces using software for manipulating 3D meshes, in this case Blender. Fig.~\ref{fig:BWLmesh}(a) shows the 3D mesh of the point cloud in Fig.~\ref{fig:BWLpre} using the Poisson Reconstruction algorithm. The  orange borders indicate adjacent vertices that should not have been connected. If the 3D mesh has minor errors, those can be corrected manually with results in Fig.~\ref{fig:BWLmesh}(b). However, if
the mesh is notably erroneous, it may be worthwhile to make sure that the data has been properly pre-processed or to apply another reconstruction algorithm. For example, if the latitude and longitude are not converted from the WGS84 system to the UTM coordinate system, the resulting mesh will be a single beam-like structure with points at different depths looking as if they are super-imposed.



\subsection{Generating the Above-Waterline 3D Mesh}

\begin{figure}[h]
\centering
\includegraphics[width=0.3\textwidth]{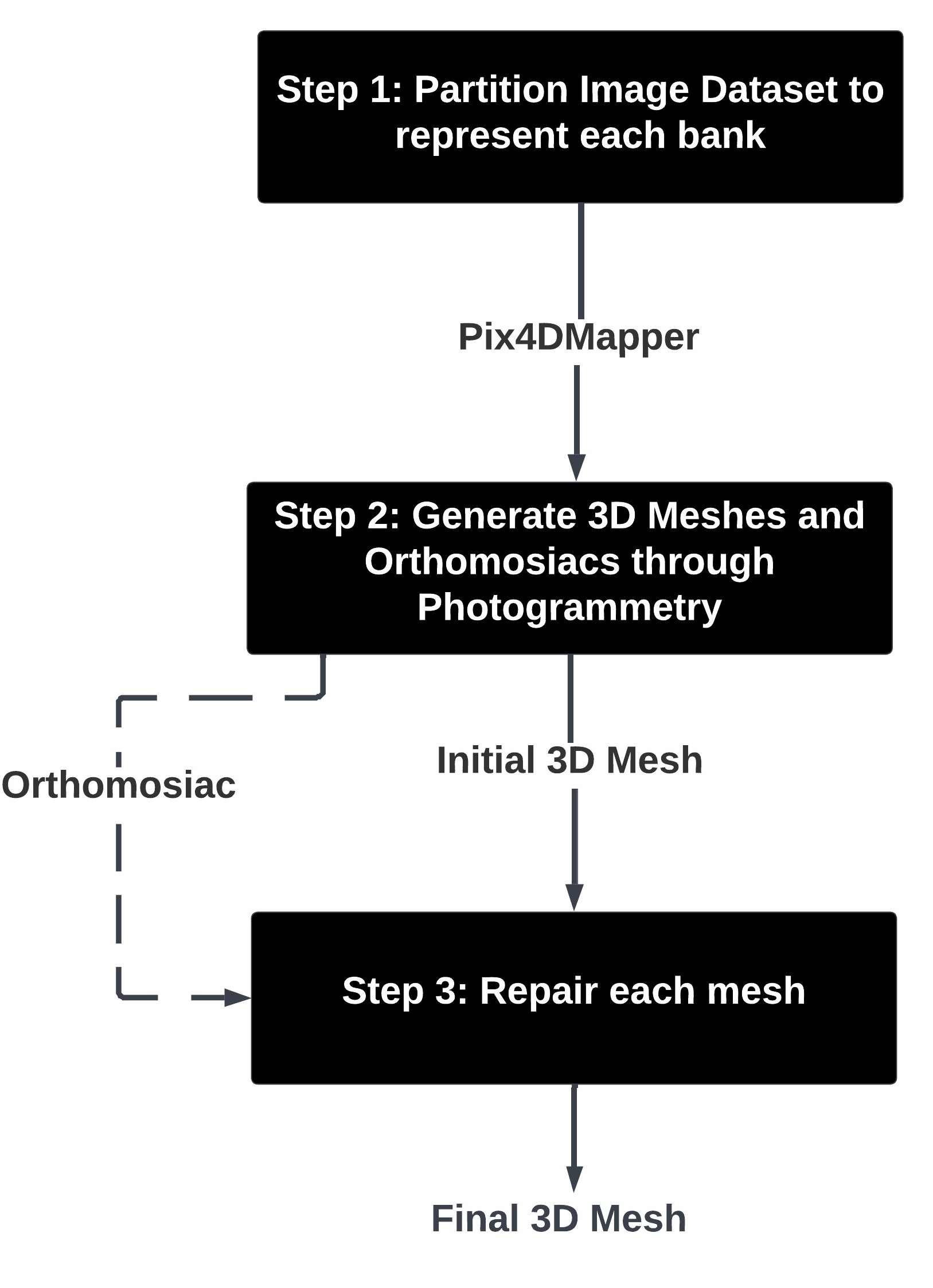}
\caption{Steps in generating the Above-Waterline mesh.}
\label{fig:generateAbove}
\end{figure}

Fig.~\ref{fig:generateAbove} summarizes the steps in generating the above-waterline mesh. 
Step 1 is to partition the collected images. 
The above-waterline dataset contains a densely populated dataset of the  banks of the body of water. The images are partitioned into different banks, either left or right of a river or roughly linear stretches of a pond.
Partitioning reduces the work time for Pix4DMapper significantly. It also removes the complexity of dealing with curved banks. 
This allows for the processing of each bank in a different instance of the software and then stitching all the above-waterline meshes in post-processing. 

Step 2 generates the mesh for each partitioned dataset. It should be noted that each image should be geotagged in order to reliably produce an accurate mesh. Although the approach does not depend on specific photogrammetric packages, Pix4DMapper was used. For the sake of replicability, the settings for
the Pix4DMapper are given here: 
 Geolocation Accuracy should be Standard; the Processing Options Template should use 3D Models; and Processing should have Initial Processing; Point Cloud and Mesh; DSM, Orthomosaic, and Index checkboxes checked. 
If there is GPS data associated with the images, each image will be shown as a data point on the Map View screen of the software, during the processing stage.

\begin{figure}[h]
\centering
\includegraphics[width=0.45\textwidth]{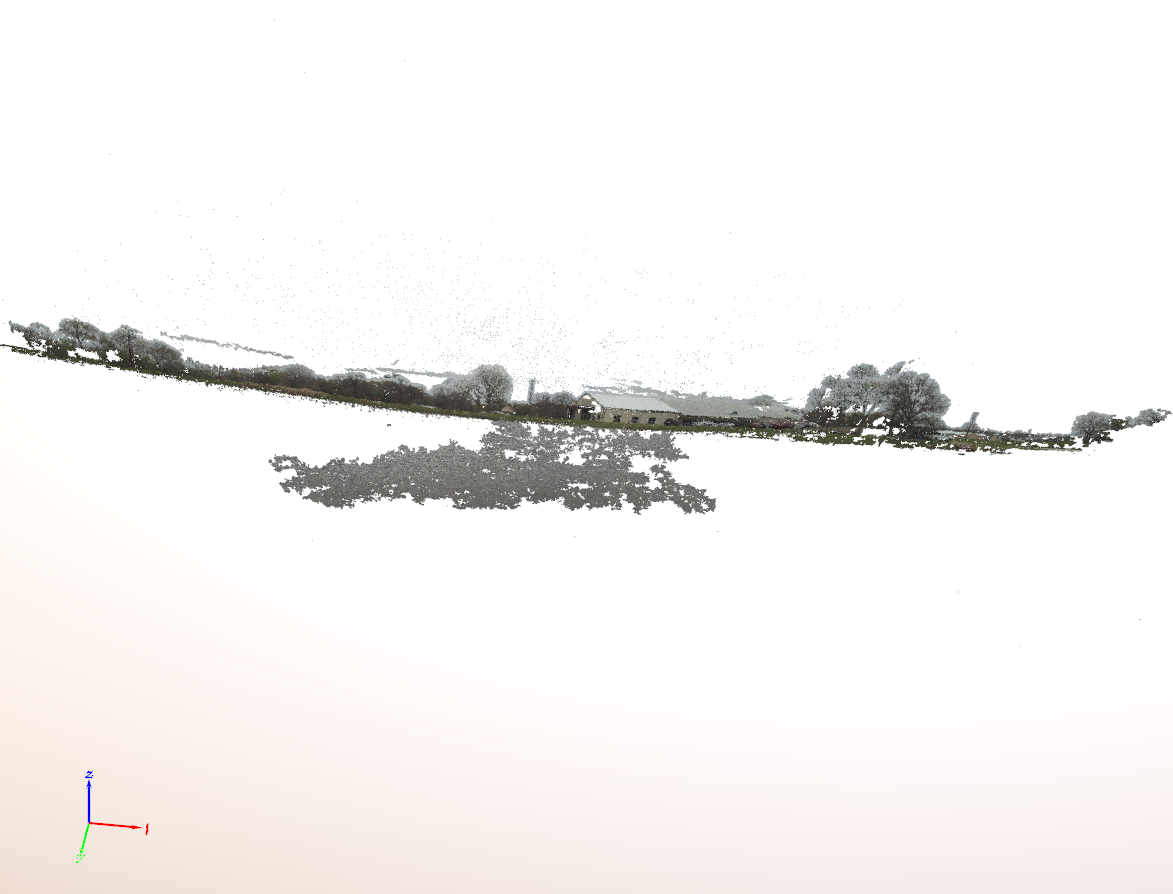}
\caption{Point cloud of a bank showing a building and reflections in the water.}
\label{fig:bank}
\end{figure}

After the initial processing in Step 2 has been completed, the software will produce a point cloud of the 3D Mesh which is representative of the final mesh that is going to be created. An example is shown in Fig.~\ref{fig:bank}, with a bank in front of a building. Note the artifacts from reflections in the water below the bank.

Step 3 is to repair the mesh. As seen in Fig.~\ref{fig:bank}, the photogrammetric package may recognize water surface and cloud surface as points of interest, and create vertices correlated to the clouds and water surface. 
As with Step 3 in the below-waterline process, it is helpful to use the point clouds from Step 2  to identify the vertices that are not representative of the bank, and correspondingly remove them from the mesh. 

\subsection{Merging the Below- and Above-Waterline Meshes}\label{alignment}

\begin{figure}[h]
\centering
\includegraphics[width=0.45\textwidth]{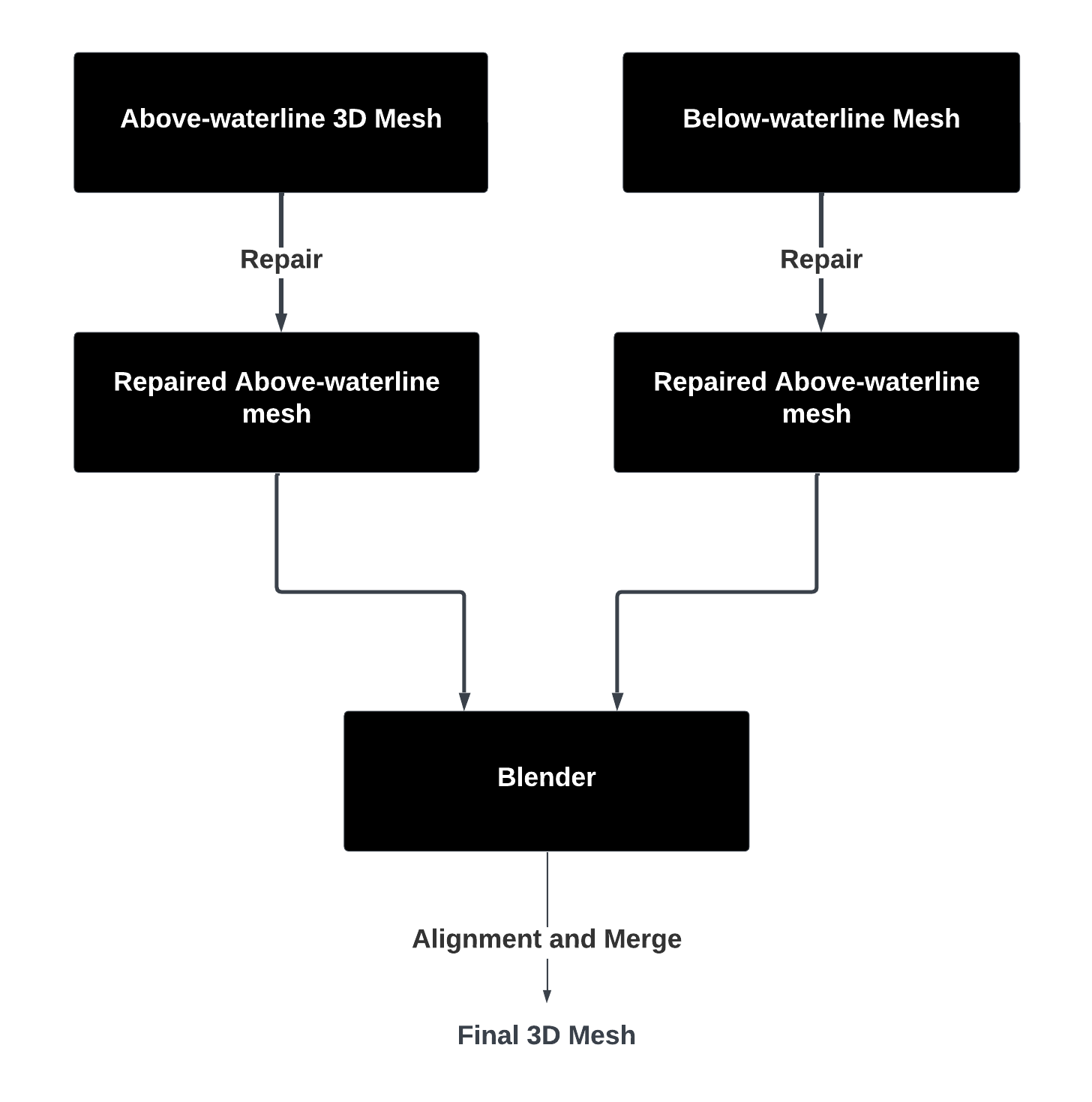}
\caption{Overview of How to Generate the Final Mesh}
\label{fig:OverviewMerge}
\end{figure}

Once all the meshes have been individually repaired, the meshes can be aligned and merged using the Blender software package following Fig.~\ref{fig:OverviewMerge}. Blender produces a resulting point cloud and a mesh that can be exported in the .PLY or .OBJ File Format. The merged point cloud can be used to check for errors. 

In order for Blender to merge the meshes, all meshes should be georeferenced in the same
coordinate system. It is possible to drag-and-drop meshes to manually align them, but having the meshes geolocated automates this. If the above-waterline meshes are created with Pix4DMapper, those meshes will require intermediary steps to geolocate them in the WGS84 coordinate system. These intermediary steps consist of extracting  
 the offset information capturing the distance from the GPS location of the camera found in the Pix4DMapper project folder, 
adding the offset to each vertex of the corresponding above-waterline 3D Mesh, and then
    converting the coordinates of each vertex from UTM to WGS84.

\section{Demonstration: Lake ESTI}



\begin{figure}[!tbp]
  \centering
  \subfloat[Polygon of area surveyed]{\includegraphics[width=0.45\textwidth]{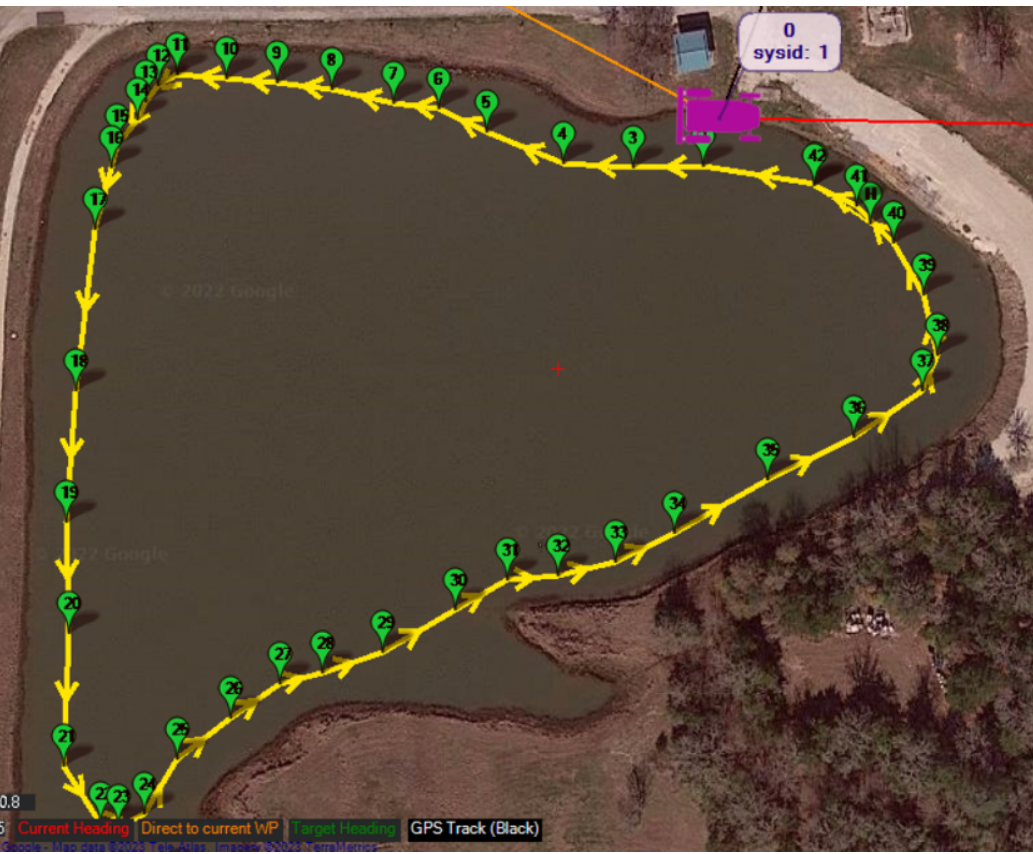}}
  \hfill
  \subfloat[Approximate Image Location of the Banks]{\includegraphics[width=0.45\textwidth]{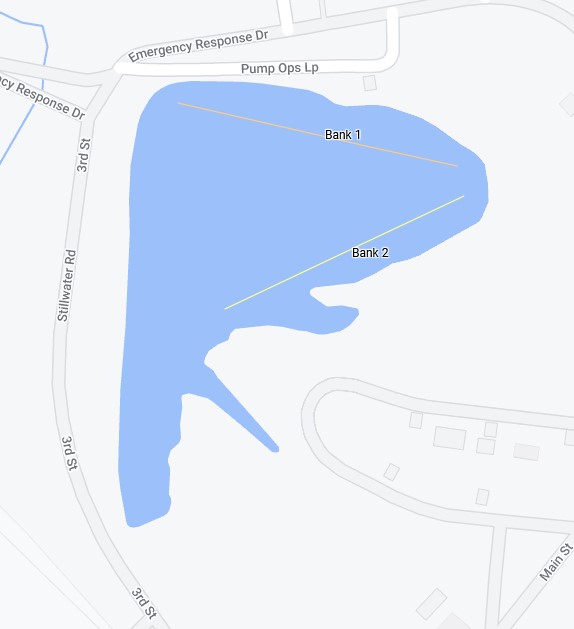}}
  \caption{Data collection at Lake ESTI.}
  \label{fig:banks}
\end{figure}

In order to demonstrate this approach, data was collected using a Hydronalix EMILY USV with a Humminbird single beam echosounder and a Teledyne FLIR Duo Pro R camera at Lake ESTI,  a  $18,580m^2$ naturally occurring pond at the Texas A\&M Engineering Extension Service Disaster City\textsuperscript{\textregistered} complex. Lake ESTI has a triangular shape with three well-defined banks, with one bank covered with trees, another side a flat field, and the third, a levee and a building. Prior work had established the accuracy of the  bathymetric map generated by the echosounder for Lake ESTI  and structure from motion for above-waterline mapping is well known. Thus, the purpose of the field exercise was not to show accuracy but rather to illustrate the workflow, and challenges, in collecting the data from the USV. 

Fig.~\ref{fig:banks}(a) shows the outline of the polygon bounding the survey area superimposed on a satellite image of Lake ESTI. The boundary was set to
cover the pond without the USV running aground; note this left a gap between the bank and the boat. 
EMILY planned and autonomously executed a complete coverage path \cite{trey} collecting underwater depth data. The path required multiple batteries to be swapped out resulting in the polygon being covered in three sections. At the completion of each section, EMILY then followed the bank side of the polygon and collected camera images for the above-waterline mesh.

Unfortunately, it was discovered in post-processing that the camera and RTK readings used to geotag camera images were only intermittently recorded for the first section, though apparently the problem resolved itself after the battery change and it worked correctly for the remaining two sections. This meant that the above-waterline mesh could not be successfully constructed for the first section. The echosounder data was still usable as it had a built-in GPS system,  so the bathymetric map was complete for all three sections of the polygon. Fig.~\ref{fig:banks}(b) indicates the two banks that were correctly captured.






\begin{figure}[!tbp]
  \centering
  \subfloat[Bank 1 in upper right corner.]{\includegraphics[width=0.45\textwidth]{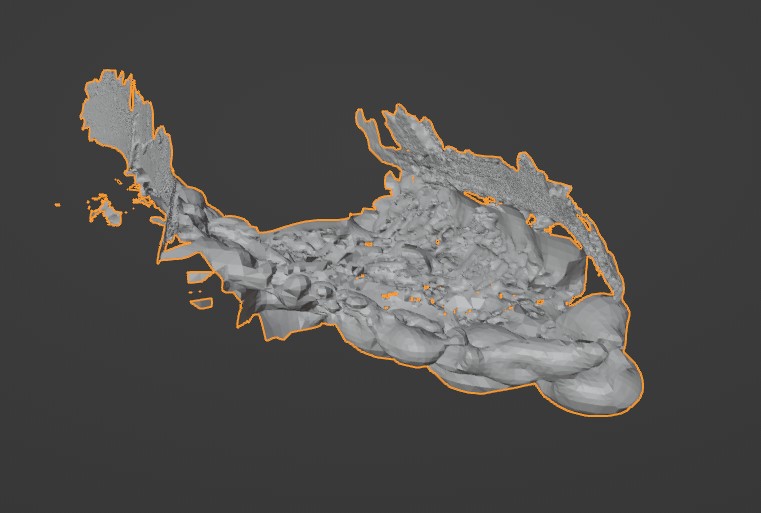}\label{fig:f3}}
  \hfill
  \subfloat[Bank 2 in upper right corner]{\includegraphics[width=0.45\textwidth]{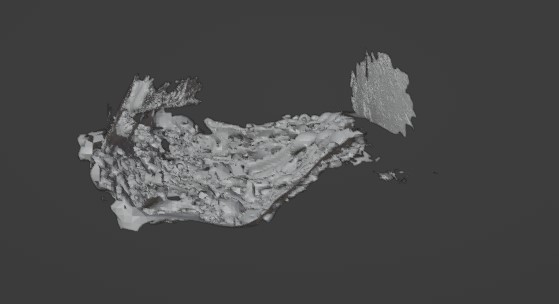}\label{fig:f4}}
  \caption{Two rotations of the final render of Lake ESTI.}
  \label{fig:blender}
\end{figure}

Fig.~\ref{fig:blender} shows two rotations of the final combination of the three meshes (below-waterline, bank 1, bank 2). Notice the gaps between the below and above the waterline mesh in Fig.~\ref{fig:blender}(b). 



\section{Conclusions}

This project confirms that volumetric models of non-navigable rivers and shallow bodies of water can be created from sonar and camera data captured by a small uncrewed surface vehicle. The below-waterline bathymetry and the above-waterline point cloud can be converted to georeferenced meshes and merged using the popular Blender software.  It should be noted that successfully aligning meshes without manually translating or rotating meshes to compensate for obvious errors requires 
the GPS errors between sensors to be minimized. Therefore it is recommended that both the echomapper and camera system use a common RTK system to guarantee congruence. 

The disadvantage of the approach taken by this paper is that it may require manual intervention to complete the surface for volumetric analyses during the repair steps. 
An alternative method is to turn the mesh into voxels and artificially fill up the hole by increasing the size of the voxels. Once the voxels are large enough, the mesh can be wrapped with an outer layer, essentially a surface smoothing. 
The expectation is that the mesh that will be produced will have the least amount of distortion as compared to the original mesh, and the holes will be covered up by the outer layer of the mesh. 

Ongoing work is targeting deployments in non-navigable rivers. In those bodies of water, it should be easier to collect high-quality above-waterline DEM because the banks are closer together than in a lake or pond.  However, it is more challenging to specify the polygon to survey because the rivers may have changed course and they usually contain navigational hazards such as fallen tree limbs,  debris, and sandbars. Work is experimenting with using drones to autonomously identify open water where the USV can navigate without running aground and generate the polygon. 



\section*{Acknowledgments}

The authors thank Clint Arnett at Disaster City and Valeria Heredia, Thomas Manzini, Itzel Rodriguez, Yashas Salankimatt, and Trey Smith for their assistance.

\bibliography{Bib/main}

\end{document}